\documentclass[a4paper, 11pt]{article}
\usepackage[T1]{fontenc}
\usepackage{times}
\usepackage{graphicx}
\usepackage{amsfonts}
\usepackage{amssymb}
\usepackage{amsmath} 
\usepackage{amsthm}
\usepackage{natbib}
\usepackage{graphicx}
\usepackage{pdfpages}
\usepackage{tabularx}
\usepackage{hyperref}

\usepackage[margin=1in]{geometry}

\usepackage{pifont}
\makeatletter
\def\blfootnote{\xdef\@thefnmark{}\@footnotetext}
\makeatother

\newcommand {\OMIT}[1]{}

\begin{document} 

%\date{}

\title{Benchmark of structured machine learning methods for microbial identification from mass-spectrometry data}
\author{K\'evin Vervier\,$^{1,2,3,4}$, Pierre Mah\'e\,$^{1}$,\\Jean-Baptiste\ Veyrieras\,$^{1}$ and Jean-Philippe Vert\,$^{2,3,4}$}%\footnote{$^{1}$ Bioinformatics Research Departement, bioM\'erieux, 69280 Marcy-l'\'Etoile, France , $^{2}$  MINES ParisTech, PSL Research University, CBIO-Centre for Computational Biology, 77300 Fontainebleau, France  , $^{3}$ Institut Curie, 75248 Paris Cedex, France, $^{4}$ INSERM U900, 75248 Paris Cedex, France}}

\maketitle

\blfootnote{\noindent $^{1}$Bioinformatics Research Departement, bioM\'erieux, 69280 Marcy-l'\'Etoile, France\\
$^{2}$MINES ParisTech, PSL Research University, CBIO-Centre for Computational Biology, 77300 Fontainebleau, France\\
$^{3}$ Institut Curie, 75248 Paris Cedex, France, \\$^{4}$ INSERM U900, 75248 Paris Cedex, France\\
$^*$ contact: \href{mailto:pierre.mahe@biomerieux.com}{pierre.mahe@biomerieux.com}}

\begin{abstract}
Microbial identification is a central issue in  microbiology, in particular in the fields of infectious diseases diagnosis and industrial quality control. The concept of species is tightly linked to the concept of biological and clinical classification where the proximity between species is generally measured in terms of evolutionary distances and/or clinical phenotypes. Surprisingly, the information provided by this well-known hierarchical structure is rarely used by machine learning-based automatic microbial identification systems. Structured machine learning methods were recently proposed for taking into account the structure embedded in a hierarchy and using it as additional \textit{a priori} information, and could therefore allow to improve microbial identification systems.

We test and compare several state-of-the-art machine learning methods for microbial identification on a new Matrix-Assisted Laser Desorption/Ionization Time-of-Flight mass spectrometry (MALDI-TOF MS) dataset. We include in the benchmark standard and structured methods, that leverage the knowledge of the underlying hierarchical structure in the learning process. Our results show that although some methods perform better than others, structured methods do not consistently perform better than their "flat" counterparts.
We postulate that this is partly due to the fact that standard methods already reach a high level of accuracy in this context, and that they mainly confuse species close to each other in the tree, a case where using the known hierarchy is not helpful. 

\end{abstract}

%%%%%%%%%%%%%%%%%%%%%%
\section{Introduction}
%%%%%%%%%%%%%%%%%%%%%%
Microbial identification is the task of determining to which species a microorganism isolated from a clinical or industrial sample belongs. It plays a central role in the diagnosis of infectious diseases and industrial quality control. In the clinical setting, identification is often the first step towards a finer characterization of the microorganism, aiming in general to establish its virulence and/or antibiotic resistance profiles, which is ultimately used by the clinician to prescribe a therapy.

Since the proof of concept of bacterial identification with MALDI-TOF MS~\citep{anhalt1975identification}, , this high-throughput technology has been improved up to a genuine paradigm breaking technology in microbiology, allowing to quickly, cheaply and efficiently characterize a microorganism~\citep{bizzini2010matrix,cherkaoui2010comparison,gaillot2011cost,tan2012prospective}.
Starting from an isolated colony of the targeted microorganism, MALDI-TOF MS provides a snapshot of its proteomic content.
Such a proteomic fingerprint is highly species specific, and can be used to identify a microorganism by matching it with a reference database of annotated fingerprints~\citep{van2012biomedical}.

At the basis of MALDI-TOF MS identification system is therefore a software component in charge of finding the closest match between the fingerprint of the unknown microorganisms and the reference fingerprints of the database.
From the data analysis perspective, this can be formalized as a multiclass classification task. 
This learning task  presents several challenging issues.
First, MALDI-TOF mass spectra are measured on several tens of thousands of mass to charge channels, and although they are generally pre-processed in order to extract their predominant peaks~\citep{coombes2007pre}, the resulting peak lists are still high-dimensional vectors.
Moreover, current commercial systems like the Biotyper (Bruker Daltonics, Germany), LT2 (Andromas, France), or VITEK-MS (bioM\'erieux, France) address several hundreds of species~\citep{Martiny2012}, which constitutes a relatively massive multiclass problem. Finally, the number of observations per class, that is, of representative strains per species, is often limited, which leads to strongly unbalanced datasets. 
On the other hand, the classes of the problem correspond to microbial species which can be organized into well known hierarchical structures,  generally defined in terms of evolutionary distances and/or phenotypic differences. 
Such tree structures provide a rich source of information that could be added as prior knowledge within the training of automatic microbial identification systems. Several "structured" machine learning methods were indeed recently proposed for taking into account the structure embedded in a hierarchy and using it as additional \textit{a priori} information~\citep{hofmann2003learning,tsochantaridis2005large,sun2001hierarchical,dumais2000hierarchical}, and could potentially be used to train microbial identification systems. Surprisingly, however, this possibility has not been investigated to our knowledge, and current systems implement "flat" multiclass classification algorithms that do not take into account the known tree structure.
In this paper, we evaluate the relevance of structured machine-learning methods in the context of microbial identification from MALDI-TOF mass spectra. For that purpose,
we use the MicroMass dataset~\citep{mahe2014automatic} to benchmark several "flat" and "structured" machine learning techniques.

%%%%%%%%%%%%%%%%%%%%%%%%%%%
\section{Material and Methods}
\subsection{Benchmark dataset}\label{chap2:micromass}
%%%%%%%%%%%%%%%%%%%%%%%%%%%

The dataset considered in this benchmark is described in Table~\ref{Tab:MicroMass}. It involves 20 Gram positive and negative bacterial species covering nine genera.
This dataset was extracted from the reference database embedded in the commercial VITEK-MS system and made public through the UCI machine learning repository\footnote{\url{http://archive.ics.uci.edu/ml/datasets/MicroMass}} .
Each species is represented by 11 to 60 mass spectra obtained from 7 to 20 bacterial strains, leading altogether to a dataset of 213 strains and 571 spectra.
These spectra were obtained according to the standard workflow used in clinical routine in which the microorganism was first grown on an agar plate from 24 to 48 hours, before some colonies were picked, spotted on a MALDI slide and a mass spectrum was acquired.

	\begin{table}[!ht]
	\caption{
	\textbf{MicroMass dataset}. This table describes the MicroMass dataset content, in terms of used bacterial genera and species. It also provides information on the number of bacterial strains and mass-spectra for each species.}
	
	\begin{tabular}{|l|c|c|c|}
	\hline
	Species name & Species ID & Number of strains & Number of spectra \\\hline
	\textit{Bacillus cereus} & BAC.CEU & 10 & 26 \\
	\textit{Bacillus thuringiensis} & BAC.THU & 8 & 11\\\hline
	\textit{Citrobacter braakii} & CIT.BRA & 9 & 26 \\
	\textit{Citrobacter freundii} & CIT.FRE & 10 & 28 \\\hline
	\textit{Clostridium difficile} & CLO.DIF & 7 & 14 \\
	\textit{Clostridium glycolicum} & CLO.GLY & 9 & 16 \\\hline
	\textit{Enterobacter asburiae} & ENT.ASB & 10 & 29 \\
	\textit{Enterobacter cloacae} & ENT.CLC & 16 & 52 \\\hline
	\textit{Escherichia coli} & ESH.COL & 20 & 60 \\\hline
	\textit{Haemophilus influenzae} & HAE.INF & 18 & 50 \\
	\textit{Haemophilus parainfluenzae} & HAE.PAR & 9 & 21\\\hline
	\textit{Listeria ivanovii} & LIS.ISI & 9 & 29 \\
	\textit{Listeria monocytogenes} & LIS.MNC & 10 & 31 \\\hline
	\textit{Shigella boydii} & SHG.BOY & 9 & 18 \\
	\textit{Shigella flexneri} & SHG.FLX & 10 & 32 \\
	\textit{Shigella sonnei} & SHG.SON & 10 & 31 \\\hline
	\textit{Streptococcus mitis} & STR.MIT & 10 & 26 \\
	\textit{Streptococcus oralis} & STR.ORA & 9 & 24 \\\hline
	\textit{Yersinia enterocolitica} & YER.ETC & 10 & 27 \\
	\textit{Yersinia frederiksenii} & YER.FRD & 10 & 20 \\\hline
	\end{tabular}
	\label{Tab:MicroMass}
\end{table}

\begin{figure*}[!tpb]
\centerline{\includegraphics[width=1.2\textwidth]{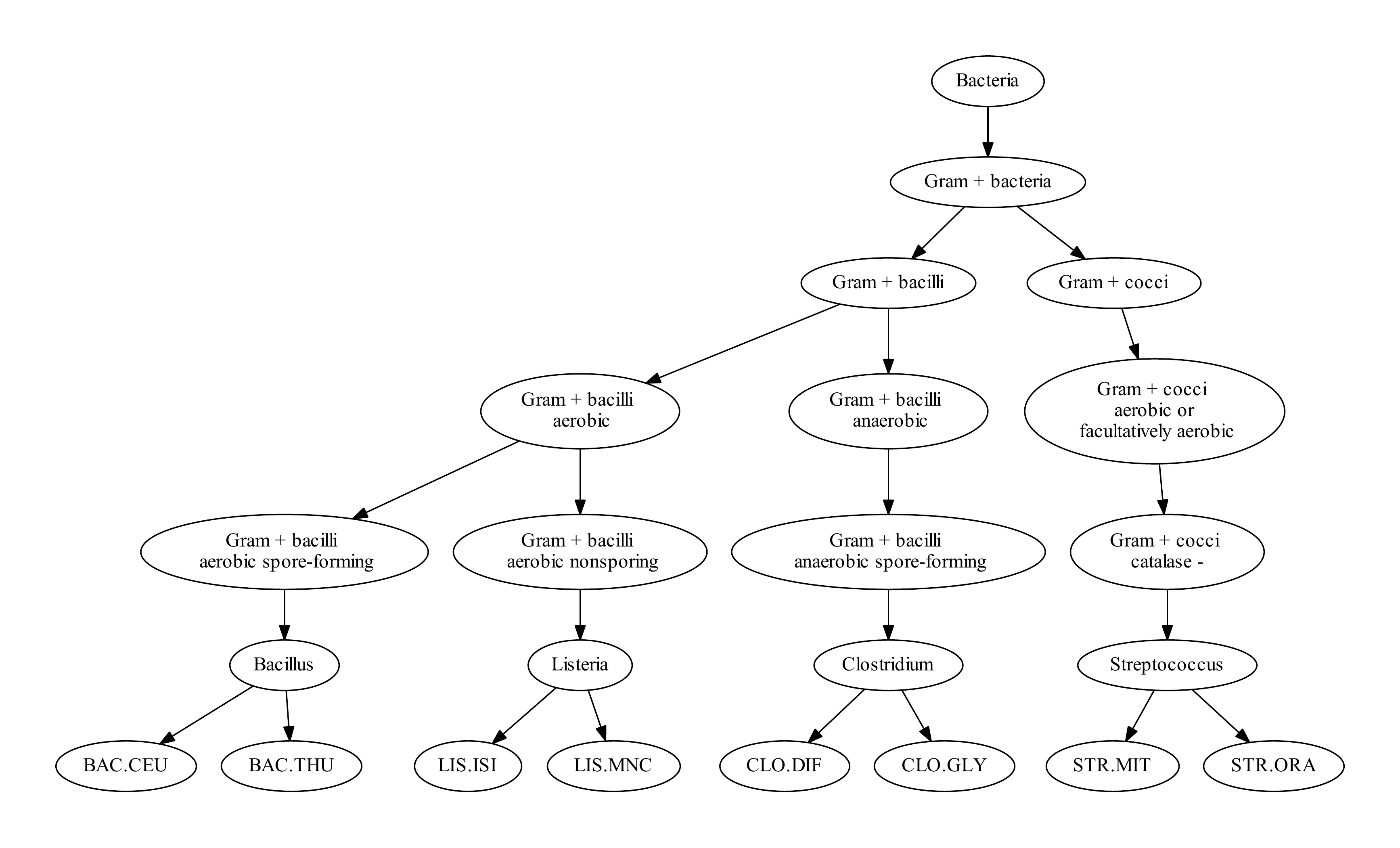}}
\caption[\textbf{MicroMass hierarchical tree structure (Gram + bacteria).}]{\textbf{MicroMass hierarchical tree structure (Gram + bacteria).}
This tree shows the hierarchical organization of the bacterial panel considered in this benchmark, that belong to the Gram + bacteria. The leaves of the tree correspond to the 8 species and their parent to the 4 genera. Internal nodes correspond to either phenotypic  ({\it e.g.} aerobic and anaerobic at the top of the tree) or taxonomic attributes.}\label{fig:taxonomy}
\end{figure*}

\begin{figure*}[!tpb]
\centerline{\includegraphics[width=1.3\textwidth ,height=9cm]{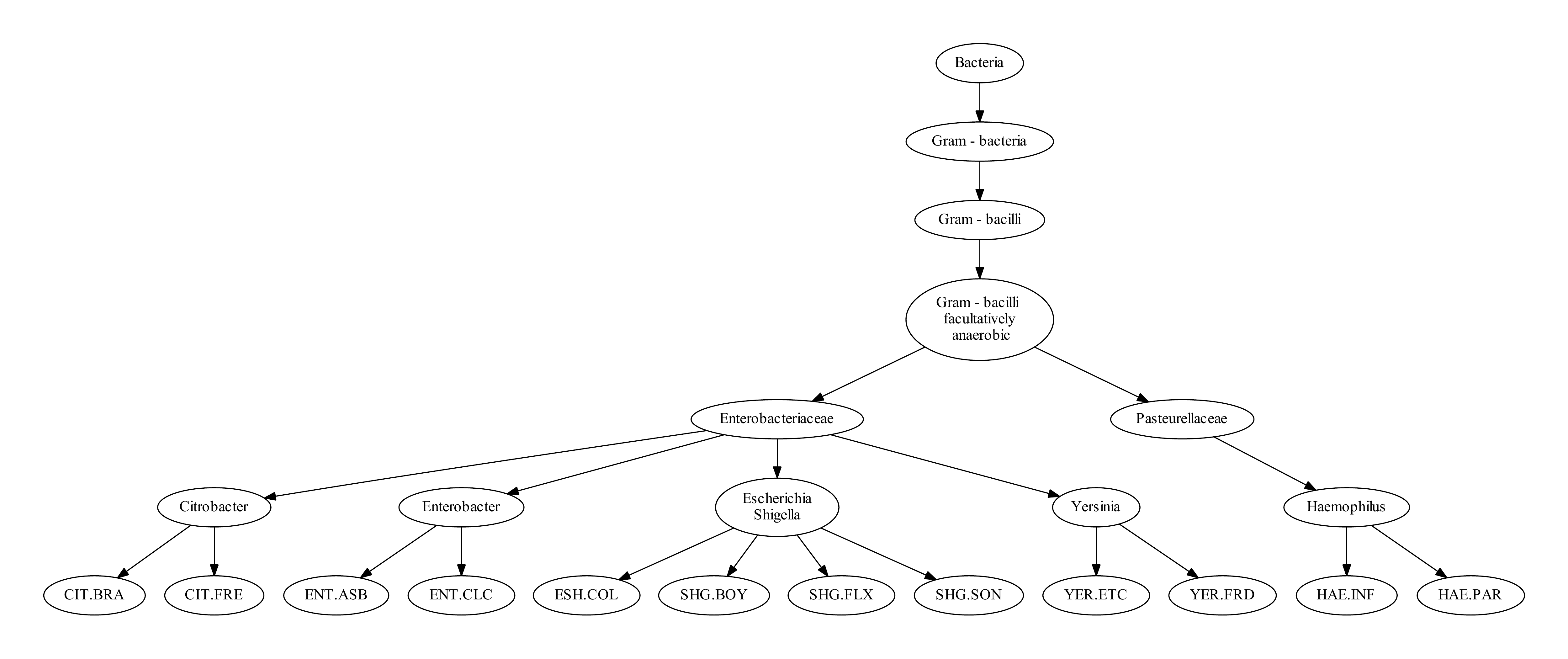}}
\caption[\textbf{MicroMass hierarchical tree structure (Gram - bacteria).}]{\textbf{MicroMass hierarchical tree structure (Gram - bacteria).}
This tree shows the hierarchical organization of the bacterial panel considered in this benchmark, that belong to the Gram - bacteria. The leaves of the tree correspond to the 12 species and their parent to the 5 genera. Internal nodes correspond to either phenotypic  ({\it e.g.} aerobic and anaerobic at the top of the tree) or taxonomic attributes.}\label{fig:taxonomy2}
\end{figure*}

The 20 bacterial species involved in this study and the underlying hierarchical tree are shown in Figures~\ref{fig:taxonomy} and \ref{fig:taxonomy2}. %, or in Appendix~\ref{Annexes:MicroMassTaxo}, Figure~\ref{fig:taxonomy3}.

%Gram positive and negative at the top of the tree)

We note that the tree considered in this study involves both phenotypic and evolutionary traits, its uppermost level separating species into Gram positive and Gram negative, and its two lowest levels corresponding to the species and genus taxonomic ranks.
Such a hybrid hierarchical definition is common  in the context of clinical microbiology, where manual identification involves a succession of tests to establish several phenotypic and metabolic properties of the microorganism to identify (e.g., Gram +/- or aerobe/anaerobe). These properties correspond to the upper levels of the tree, while the lower ones correspond to standard phylogenetic ranks (e.g., family, genus and species).
We also note that this dataset contains several pairs of groups of species known to be hard to discriminate in general. This is for instance the case of the {\it Bacillus cereus} and {\it Bacillus thuringiensis} species, which are known to belong to the {\it Bacillus cereus group} \citep{helgason2000bacillus}, as well as the {\it Escherichia coli} species  and the species of the {\it Shigella} genus, which are often considered to belong to the same species~\citep{lan2002escherichia}. Accordingly, {\it Escherichia coli} and the three {\it Shigella} species involved in this dataset  were gathered in a common genus.

We note finally that we have considered in this study a peak-list representation in which a mass spectrum is represented by a vector $x \in \mathbb{R}^p$, where $p$ is the numbers of bins considered to discretize the mass to charge range, and each entry of $x$ is derived from the intensity of the peak(s) found in the corresponding bin. 
While several schemes have been proposed to define such a peak-list representation~\citep{coombes2007pre}, we have relied here on the approach embedded in the VITEK-MS system, which provides a peak-list representation of dimension $p=1300$, with typically between 50 and 150 peaks per spectrum. 
Further details about this dataset are available in~\citep{mahe2014automatic}.

\subsection{Classification methods}\label{sec:methods}
In this section we provide a brief description of the various classification strategies considered in this study.
We refer the interested reader to the original publications for a more detailed presentation.
We first describe the standard support vector machine (SVM) algorithm and its multiclass extensions.
%We then describe three approaches to leverage information about a hierarchical structure reflecting the proximity of bacterial species: 
%\begin{itemize}
%	\item cost-sensitive multiclass SVMs, a direct extension of the multiclass SVM formulation;
%	\item hierarchical SVMs, which makes a further use of the tree structure;
%	\item and a divide-and-conquer or "cascade" strategy.
%\end{itemize}
We then describe three approaches to leverage information about a hierarchical structure reflecting the proximity of bacterial species: a cost-sensitive SVM formulation, a hierarchical SVM formulation, and a divide-and-conquer or "cascade" strategy.
Finally, we present other standard machine learning algorithms not based on SVMs, such as nearest-neighbours and random forests, that were included in the benchmark.

\paragraph{Multiclass SVMs}
In its original form, the SVM algorithm is a binary classification algorithm \citep{Cortes95}. 
It aims to build a classification rule to classify instances from the space $\mathcal{X}=\mathbb{R}^p$ into two classes, commonly referred to as positive or negative.
In its simplest linear form, the SVM algorithm builds a hyperplane separating the vector space $\mathcal{X}$ in two half-spaces.
For that purpose, the SVM algorithm relies on a training dataset of $N$ instances $x_1,\ldots, x_N \in \mathcal{X}$ with their associated labels $y_1, \ldots, y_N \in \{-1,1\}$, and seeks to correctly classify the training data while maximizing the margin of the hyperplane, which is defined as the smallest distance between the hyperplane and the training instances. 
These two criteria are hard to fulfill simultaneously, and in practice the SVM algorithm achieves a trade-off between these two objectives.
When instances must be classified into $K>2$ classes, an extension of this binary SVM is needed. The most standard way to address multiclass classification problems with SVMs is to train and combine several binary classifiers into a mutliclass classification rule. Two popular schemes allow to do so:  
\begin{itemize}
	\item the {\it one versus all} scheme (\texttt{SVM-OVA}), where a set of $K$ binary SVMs is trained to separate each of the $K$ classes from the $K-1$ other ones, leading to a set of $K$ predictors. To predict the class of a new instance $x$, each of the $K$ predictors is applied to $x$, and the one predicting with the largest margin is the winner.
	\item the {\it one versus one} scheme (\texttt{SVM-OVO}), where $K(K-1)/2$ binary SVMs are trained to distinguish between every pair of classes. The class predicted for an instance $x$ is the one obtaining the highest number of votes (a number between 0 and $K-1$) according to these classifiers. 
\end{itemize}
In a similar spirit to the one-versus-all scheme, an alternative strategy proposed by \citep{tsochantaridis2005large} and referred to as \texttt{Multiclass} below, consists in learning a set of class specific classifiers, but to combine them in a single model and to learn them simultaneously.
This formulation requires to solve a single optimization problem, but with a greater number of constraints than each individual SVM  involved in the one-versus-all formulation.
In practice, efficient algorithms enable to obtain an approximate solution of this problem \citep{tsochantaridis2005large}.
This formulation paves the way to the development of cost-sensitive and so-called structured classifiers, that we introduce in the two following sections.

	\paragraph{Cost-sensitive multiclass SVMs}\label{sec:cost-sensitive}
	%------------------------------------------
For practical applications, different errors can have different impact: it can be less severe to mistake class A for class B than class A for class Z for instance.
This is notably the case for microbial identification which can orient therapy before antibiotic susceptibility results are available.
Cost-sensitive classifiers distinguish between the various types of classification errors and penalize them differently in the learning process.
The above multiclass formulation can be easily modified to accommodate such a cost-sensitive mechanism \citep{tsochantaridis2005large}.
Indeed, assume that a loss function $\Delta : \mathcal{Y} \times \mathcal{Y} \rightarrow \mathbb{R}$ is available\footnote{Note that in practice this loss function can be summarized as a $K \times K$ matrix.} such that $\Delta(y,y')$ quantifies the loss, or severity, of predicting class $y'$ if the true class is $y$, $\Delta(y,y') > 0$ for $y\neq y'$ and $\Delta(y,y) = 0$.
Such a loss function can be leveraged in the training process through a redefinition of the constraints involved in the underlying optimization problem.
This redefinition has the effect of adjusting the strength of the constraints according to the loss function.
Note that the standard formulation corresponds to using a binary loss function: $\Delta(y,y') = 1$ for $y\neq y'$ and $\Delta(y,y) = 0$.
For practical applications, this cost-sensitive formulation allows to leverage the training process prior information about the relationship between the classes and/or requirements about the classification performances expected.
In this study we call this approach \texttt{TreeLoss} and use $\Delta(y,y')$ as the length of the shortest path connecting the two species in the considered tree. 

	\paragraph{Hierarchy structured SVMs}
	%-------------------------------------
The structured SVM formulation of \citep{tsochantaridis2005large} enables to  make a further use of the hierarchical structure underlying the microbial identification multiclass problem.
Indeed, it does not only allow to leverage a loss function in the learning process to penalize misclassifications involving hierarchically distant species, but it also introduces new variables that can be further exploited by the algorithm.
Loosely speaking, the original variables are repeated as "blocks" for each node of the tree.
These variables are then turned on or off, for each observation, depending on its position on the hierarchy : every block of variables associated to a node that belongs to the path connecting the root node to the leaf node corresponding to the observation label is turned on, and is turned off otherwise.
As a result, the closer two observations are in the tree, the more variables they share, which can be used by the algorithms to learn coarse to fine association rules \citep{hofmann2003learning,tsochantaridis2005large}.
This approach, which we refer to as \texttt{Structured} below, and the cost-sensitive multiclass formulation introduced above have in common to leverage a loss function to take into account the severity of the classification errors. 
This property can be expected to increase the quality of the predictions made by these algorithms, which will be trained to avoid "severe", that is, high loss, classification errors.
As in the previous paragraph, we define $\Delta(y,y')$ as the length of the shortest path connecting nodes $y$ and $y'$ in the tree, hence directly define the notion of severity as the tree distance.

	\paragraph{Cascade approach}
	%----------------------------
The last SVM-based strategy we consider is a divide and conquer approach where a SVM classifier is learned at each internal node of the tree to assign a spectrum to one of its children.
A top-down approach is then used to classify a spectrum to a leaf node by this cascade of classifiers~\citep{sun2001hierarchical,dumais2000hierarchical}.
Although any type of classifier can be considered at each node, we choose to rely on SVM in this study and call the resulting method {\it Cascade-of-Classifiers} (\texttt{CoC}).
Finally, following~\citep{Benabdeslem06}, we consider a variant of this approach in which the tree used to define the cascade is obtained in a preliminary step of unsupervised clustering carried out from species-specific prototypes.  
We refer to this approach as {\it Dendrogram-SVMs} (\texttt{DSVM}). % \citep{everitt2001cluster}.

%Please note that DSVM and CoC were implemented in \verb|R|, with \verb|LiblineaR| SVM at each node.

	\paragraph{Other classification methods}
	%-------------------------------------
Finally, we consider three methods not based on SVMs in this benchmark. Random forest (RF) and similarity-based approaches have indeed already been successfully used in the context of MS data classification \citep{Debruyne11,Taskin13}. We therefore include in the benchmark the RF method described in \citep{Breiman01}, referred to as \texttt{RF}, which consists in learning many decision trees and predicting with a majority vote strategy. We also evaluate two similarity-based approaches: a \textit{1-nearest-neighbour} (\texttt{1-NN}) and a \textit{1-nearest-centroid} (\texttt{1-Centroid}) approach. In the 1-NN method, a new spectrum is classified in the same class  as its closest spectrum in the training set. 
The same classification rule is applied in the nearest-centroid approach, the centroid of a given species being defined as its median spectrum.

\subsection*{Experimental setting}\label{sec:settings}

We evaluate the classification performance of the various methods by cross-validation.
To define the cross-validation folds we take into account the strain information.
Indeed, the dataset consists of 571 spectra obtained from 213 strains, with in average less than 3 and up to 6 spectra per strain.
The variability observed within the replicate spectra of a given strain is purely technical, and is therefore lower than the level of variability that is expected in clinical routine, where an additional level of biological variability is expected due to the fact that the microorganisms to identify differ from that used to learn the classification model.
To mimic this setting, hence to avoid optimistic evaluation of classification performance, we therefore affect spectra of a given strain to the same cross-validation fold.
In this study, we actually resort to a {\it leave one strain out} cross-validation strategy in which a single strain is kept aside at each step, thus leading to a 213-fold cross-validation set up.

To assess the classification performance, we primarily consider an accuracy criterion.
However, since each species of the dataset is represented by a varying number of strains and each strain by a varying number of spectra, we adopt a nested definition of accuracy criterion, instead of classical proportion of correct classifications. We first define a {\it strain-level accuracy} as the proportion of spectra that are correctly classified for each strain, and a {\it species-level accuracy} as the average strain-level accuracy for each species.
The overall accuracy indicator is then defined as the average species-level accuracy. In order to compare the benchmarked approaches, we rely on the two-sample Kolgomorov-Smirnov test \citep{kolgomorov33,smirnoff39} applied on vectors of 20 accuracies at the species level. %\textit{paired t-test} \citep{Student08}
%KOLMOGOROV-SMIRNOV TEST INSTEAD, SAME RESULTS, BUT NO NORMALITY ASSUMPTION !!!!
 %computed on the species-level accuracy vectors of size 20.
The second performance indicator we consider is the distribution of the tree loss of misclassifications, which therefore quantifies their severity.
As can be read from Figure \ref{fig:taxonomy}, this loss can vary in this study from 2 to 12, when a species is respectively mistaken for a species of the same genus or of the other Gram.
Because these types of errors are easier to interpret than summary statistics of the tree loss distribution, we report the proportion of errors that fall in the following categories: "within-genus error" ($\Delta = 2$), "outside genus but same Gram error" ($2<\Delta<12$), "distinct-Gram error" ($\Delta = 12$).

The regularization ($C$) parameter of the various SVM formulations considered in this study was optimized within each fold of the leave one strain out cross-validation process by an inner 10 fold cross-validation.
As before, spectra of the same strain are systematically affected to the same fold.
The grid of candidate values was set to $\{10^{-6},10^{-2},...,10^2,10^6\}$ and the value was chosen to maximize the nested accuracy indicator defined above. 
The standard and cascade SVM approaches (\texttt{SVM-OVA}, \texttt{SVM-OVO}, \texttt{CoC} and \texttt{DSVM}) were implemented using the \verb|R| package \verb|LiblineaR|\footnote{\url{http://cran.r-project.org/web/packages/LiblineaR/}}. 
For the two cascade approaches (\texttt{CoC} and \texttt{DSVM}), one-versus-all classifiers were trained at each internal node of the hierarchy.
The tree involved in the \texttt{DSVM} method was generated by the \verb|hclust| function of the \verb|R| package \verb|stats|, with a complete linkage clustering method. 
The \texttt{Multiclass} SVM implementation relies on the \verb|C| library \verb|SVM-light| \citep{Joachims1999}.
The cost-sensitive (\texttt{TreeLoss}) and \texttt{Structured} SVM formulations were implemented based on the \verb|C| library \verb|SVM-struct| \citep{Joachims2009}.
We have relied on the slack-rescaling approach to integrate the loss function $\Delta$ in the learning process. We have moreover considered a precision of $\epsilon=0.1$ on the solution and used the 1-slack algorithm operating in the dual (option \texttt{w=3}).

The hyperparameters of the alternative strategies were set from preliminary experiments. 
We relied on the \verb|R| package \verb|randomForest| to build \texttt{RF} models. The number of trees ($\text{ntree}$) and variables per tree ($\text{mtry}$) of the random forest were respectively set to the default value of 500 and to 36, according to the standard heuristics $\text{mtry} = \sqrt{p}$.
Preliminary experiments revealed that these parameters had little influence on the results as long as they were sufficiently high, especially $\text{ntree}$.
Regarding similarity-based methods, the number of neighbours to consider in the nearest neighbours was set to 1 and the Euclidean distance was used.
The choice of the distance criterion had little influence on the results but performance decreased when the number of neighbours increased.
The Euclidean distance was used for the nearest centroid approach as well.

We note finally that the feature vectors were systematically scaled to unit Euclidean norm.

% Results and Discussion can be combined.
\section*{Results and Discussion}
\begin{table}[!tpb]
\centering
\begin{tabular}{lccccc}\hline
method & accuracy &\# correct &\# within-genus &\# within-Gram & \# distinct-Gram \\\hline
1-NN & 76.8 & 442 & 119 & 6 & 4 \\
1-Centroid  & 78.8 & 445 & 104 & 7 & 15\\
RF  & 84.0 & 494 & 63 & 12 & 2 \\
SVM-OVO & 86.6 & 506 & 52 & 13 & 0 \\
SVM-OVA  & 88.9 & 514 & 50 & 4 & 3 \\
Multiclass & 88.9 & 516 & 47 & 4 & 4 \\\hline
Treeloss & 89.3 & 517 & 47 & 3 & 4 \\
Structured  & 89.4 & 517 & 47 & 4 & 3 \\
CoC & 88.6 & 505 & 55 & 11 & 0 \\
DSVM & 87.1 & 507 & 56 & 2 & 6 \\\hline
\end{tabular}
\caption[\textbf{Cross-validation results on MicroMass dataset.}]{\textbf{Cross-validation results on MicroMass dataset}. This table summarizes the cross-validation results obtained for each benchmarked method.
The accuracy measure corresponds to the nested accuracy definition. The four following figures explicitly give the numbers of correct prediction, of {\it within-genus} errors (for which a species was mistaken for a species of the same genus), of {\it within-Gram} errors (for which a species was mistaken for a species of another genus of the same Gram) and of {\it distinct-Gram} errors (for which a species was mistaken for another species of the other Gram).  Method names are specified in the main text of section \ref{sec:methods}.}\label{Tab:03}
\end{table}

The results of the benchmark experiment described in the previous sections are summarized in Table \ref{Tab:03}.
Considering the overall accuracy obtained by the various methods, we first note that SVM classifiers, with an accuracy ranging from 86.6 to 89.4\%, outperform random forests (accuracy of 84\%) and similarity-based approaches (accuracy of 76.8\% and 78.8\% for the nearest neighbour and nearest centroid approaches respectively). In both cases, these differences are significant ($P$-value $ < 0.05$). Among the different SVM formulations, we see that the best structured SVM (\texttt{Structured}), with an accuracy of 89.4\%, outperforms the best "flat" SVMs (\texttt{SVM-OVA} and \texttt{Multiclass}), which reach an accuracy of 88.9\%. This difference, however, is not significant ($P$-value $> 0.05$), suggesting that the more elaborate structured SVMs are not particularly useful for this application.

This being said, a closer look at the nature of the misclassifications, given in Table \ref{Tab:03}, reveals some slight differences between the various SVM strategies.
We note indeed that while \texttt{SVM-OVA}, \texttt{Multiclass}, \texttt{TreeLoss} and \texttt{Structured} make fewer errors than \texttt{SVM-OVO} and \texttt{CoC} (54 to 57 {\it versus} 65 to 66), some of these errors involve mistaking a species for a species of the other Gram, which never occur with \texttt{SVM-OVO} and \texttt{CoC}.
This however comes at the price of an increased proportion of errors involving mistaking a species for another one of the same Gram but of another genus, and therefore suggests that a trade-off between the number and the severity of the classification errors can be achieved.
In a similar spirit, we observe some discrepancies between the results provided by the two cascade approaches (\texttt{CoC} and \texttt{DSVM}): while the two approaches lead to a similar number of classification errors, \texttt{DSVM} leads to a higher rate of uncorrect Gram errors for a lower rate of distinct genus but same Gram errors.
These two methods only differ in the tree considered, which therefore suggests that its structure has indeed an important role in the learning process and that it could be optimized \citep{song2007dependence}.

Finally, a striking observation that can be made from Table \ref{Tab:03} is that the great majority of errors involve predicting a species for a species of the same genus, for any considered method.
While this makes sense from a biological point of view, this raises at least two hypotheses to explain why the structured methods considered in this benchmark, and in particular those derived from the structured SVM formalism (\texttt{TreeLoss} and \texttt{Structured}), did not bring any improvement over their "flat" counterparts.
First, we note that with a loss function $\Delta(y,y')$ defined as the length of the shortest path between species $y$ and  $y'$ in the tree, this type of error is the less penalized one.
While this is indeed a natural and relevant definition, it can mainly be expected to limit the number of errors involving remote pairs of species, and hardly to improve over a "flat" strategy that does a limited number of errors of this kind, as this is the case for \texttt{SVM-OVA} for instance.
On the other hand, it may also be the case that the tree considered in this study is not informative below the genus level. As mentioned previously, the dataset considered in this study involves several pairs or groups of species that are known to be hard to discriminate  in general, and by MALDI-TOF MS in particular.

\begin{figure*}[!tpb]
\centerline{\includegraphics[width =0.75\textwidth ,page=21]{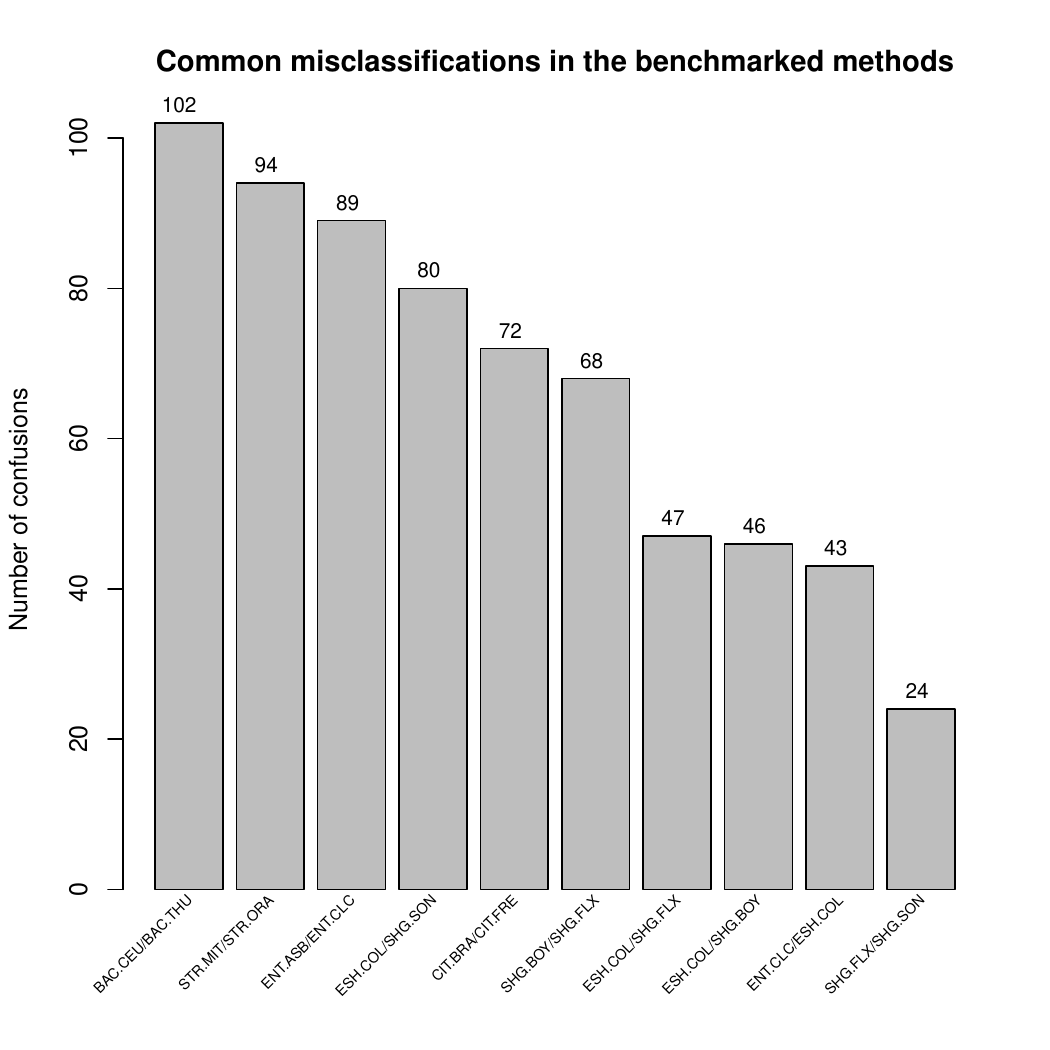}}
\caption[{\bf MicroMass dataset: Common classification errors.}]{{\bf MicroMass dataset: Common classification errors.}
Each bar represents one of the most frequent confusions observed across all evaluated classification approaches.} \label{fig:commonMistakes}
\end{figure*}

Figure \ref{fig:commonMistakes} shows the counts of the  most common types of misclassifications obtained across all the methods considered.
It reveals that five  pairs or groups of species proved to be particularly challenging: {\it Bacillus cereus} / {\it B. thuringiensis}, {\it Streptococcus mitis} / {\it S. oralis}, {\it Enterobacter asburiae} / {\it E. cloacae}, {\it Citrobacter braakii} / {\it C. freundii}, and the group defined by {\it E. coli} and the three {\it Shigella} species. It also shows that {\it E. coli} and {\it Enterobacter cloacae}, that do not belong to the same genus but both to the {\it Enterobacteriaceae} family, are relatively often mistaken. 
The biological proximity within some of these pairs or groups of species may in fact be beyond what can be captured by the MALDI-TOF technology.
The \textit{B. cereus} and \textit{B. thuringiensis} species are for instance known to belong to the \textit{Bacillus cereus} group, which is sometimes considered to define a single species \citep{helgason2000bacillus} and other studies indeed suggest that they cannot be discriminated by MALDI-TOF \citep{lasch2009identification}. \textit{Streptococcus mitis} and \textit{Streptococcus oralis} are also part of similar group comprising more than 99$\%$ 16S rRNA similarity \citep{kawamura1995determination}, and MALDI-TOF mass-spectrometry is known to be hardly able to distinguish them properly \citep{williamson2008differentiation}.

\begin{figure*}[!tpb]
\centerline{\includegraphics[width =0.9\textwidth ,page=2]{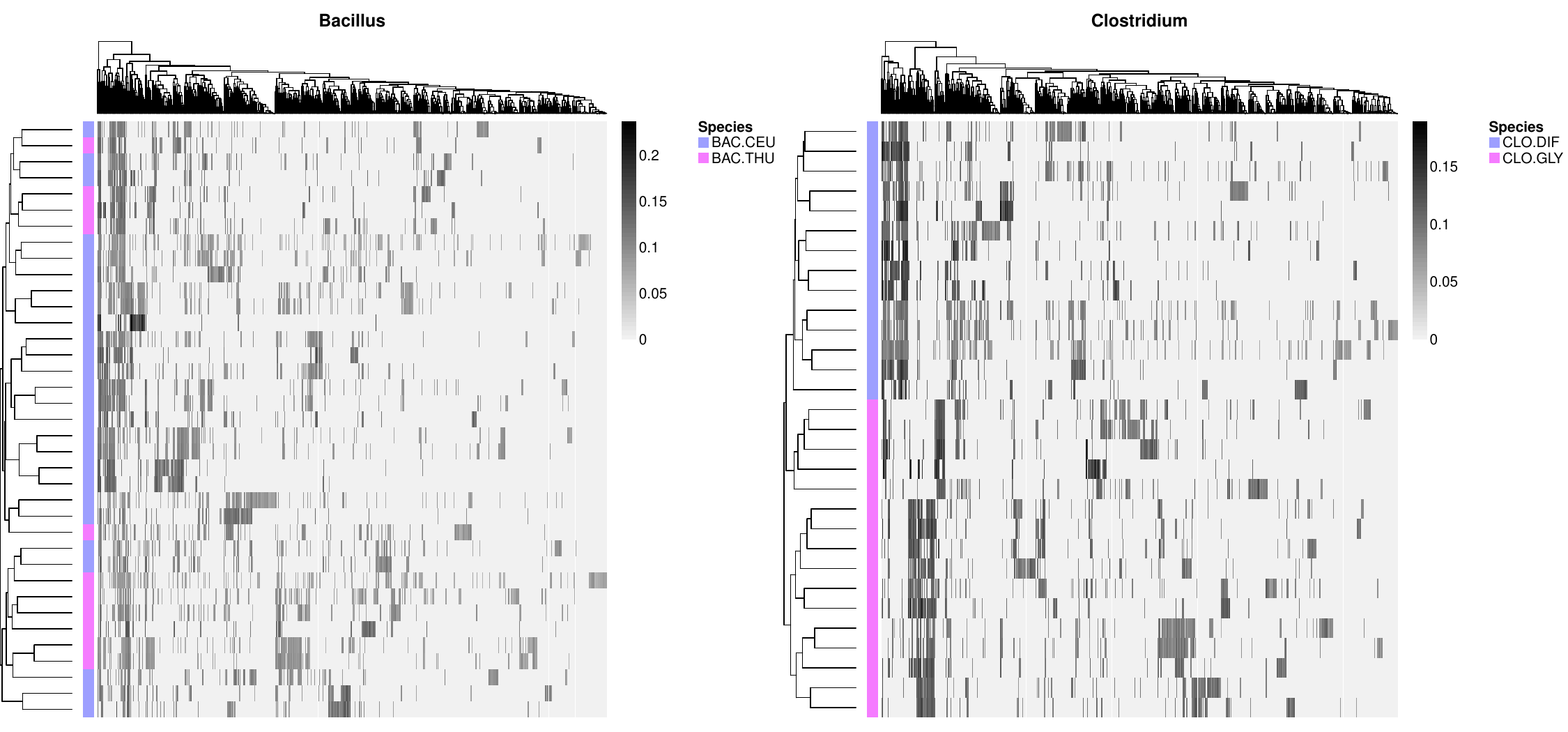}}
\caption[{\bf MicroMass dataset: Mass-spectra clustering at the genus level.}]{{\bf MicroMass dataset: Mass-spectra clustering at the genus level.} Left: \textit{Bacillus} genus. Right: \textit{Clostridium} genus. Mass-spectra (rows) belonging to a given genus are clustered according to their peak lists (columns). For clarity purpose, we removed features equal to zero among all the genus mass-spectra.
} \label{fig:bacillus}
\end{figure*}

Figure~\ref{fig:bacillus} illustrates the fact that mass spectra obtained in this study from \textit{B. cereus} and \textit{B. thuringiensis} are hardly distinguishable, at least when they have undergone the process of peak extraction, as opposed to the spectra obtained from {\it Clostridium difficile} and  {\it C. glycolicum}, that are almost never mistaken one for the other.

\section*{Conclusion}
We evaluated several structured methods in the microbial identification context, using mass-spectrometry data. 
Our results suggest that methods exploiting the underlying bacterial hierarchical structure perform as well as standard "flat" methods. 
We noted in particular that the majority of classification errors obtained by all the  methods considered in this benchmark are within-genus misidentifications.
We postulate that the structured methods considered in this benchmark are not tailored to improve flat methods for this type of errors.
Unfortunately, a larger panel of strains with a careful definition of the reference identification would be required to validate this hypothesis.
\citep{Zhou11} recently proposed a structured regularization method specifically designed to cope with this issue, and it would therefore be interesting to evaluate its relevance  in this context.

\section*{Acknowledgments}
We thank Maud Arsac for providing access to the dataset used in this study.

\end{document}